\documentclass[10pt,twocolumn,letterpaper]{article}

\usepackage[final,pagenumbers]{cvpr}
\usepackage{microtype}
\usepackage{enumitem}
\usepackage{multirow}
\usepackage{mathtools}
\usepackage{bm}
\usepackage{comment}

\newcommand{\method}{ReflexTrack\xspace}

\DeclareMathOperator*{\argmaxop}{arg\,max}

\usepackage[hidelinks]{hyperref}

\title{ReflexTrack: A Feedback-Driven Agent for Training-Free Referring Video Object Segmentation}

\author{Yuanjia Li,
        Tianyang Xu,
        Tao Zhou,
        Zhangyong Tang,
        Xiao-Jun~Wu\thanks{Yuanjia Li, Tianyang Xu, Tao Zhou, Zhangyong Tang, and Xiao-Jun Wu (\textsl{Corresponding author}) are with the School of Artificial Intelligence and Computer Science, Jiangnan University, Wuxi, China. (e-mail: yuanjia\_li133@163.com, tianyang.xu@jiangnan.edu.cn, taozhou@jiangnan.edu.cn, zhangyong\_tang\_jnu@163.com, wu\_xiaojun@jiangnan.edu.cn).},
        and Josef~Kittler\thanks{Josef Kittler is with the Centre for Vision, Speech and Signal Processing, University of Surrey, Guildford, GU2 7XH, UK. (e-mail: j.kittler@surrey.ac.uk).},
}

\begin{document}
\pagestyle{plain}
\maketitle
\thispagestyle{plain}

\begin{abstract}
Referring video object segmentation (RVOS) requires segmenting a target specified by natural language throughout a video. Recent agentic approaches combine multimodal large language models with promptable segmentation models to perform RVOS without task-specific training. However, most pipelines rely on one-shot spatial grounding followed by mask propagation, leaving both the initial prompts and temporal predictions largely unverified. We introduce ReflexTrack, a training-free, feedback-driven agent that closes this loop at both spatial and temporal levels. Mask-guided Spatial Refinement evaluates the mask induced by the current keyframe prompt and iteratively updates the bounding box together with positive and negative points, yielding a more reliable initialization. Video-level Mask Reflection assesses the complete mask sequence, localizes unreliable intervals, selects complementary repair keyframes, and generates candidate predictions through mask-guided re-propagation. Only candidates that provide a verified improvement are used to update the affected intervals, preserving reliable predictions elsewhere. All components remain frozen during inference. ReflexTrack achieves an overall $\mathcal{Q}$ score of $69.7$ on Ref-VPS and a $\mathcal{J}\&\mathcal{F}$ score of $67.2$ on ReasonVOS. These results demonstrate that prediction-level feedback substantially improves the reliability of training-free RVOS.
\end{abstract}

\section{Introduction}
\label{sec:introduction}

Referring video object segmentation (RVOS) requires segmenting a target specified by a natural-language expression throughout a video~\cite{gavrilyuk2018actor,seo2020urvos}. Beyond recognizing appearance, a model must resolve actions, spatial relations, interactions, and high-level semantics, while maintaining accurate masks under occlusion, fast motion, deformation, and visually similar distractors~\cite{ding2023mevis,yan2024visa}. These coupled language, spatial, and temporal challenges make reliable target grounding and propagation central to RVOS.

Most existing methods learn cross-modal video-text representations with supervised architectures. Transformer-based approaches formulate language as object queries and jointly perform target localization and mask decoding~\cite{botach2022mttr,wu2022referformer,yan2024mutr}, while recent methods further exploit motion-aware features, semantic-temporal alignment, global-local reasoning, and visual grounding foundations~\cite{lin2025glus,liang2025referdino}. Although effective, they typically rely on large-scale pixel-level supervision and architecture-specific training, which limits their ability to immediately benefit from rapidly evolving multimodal large language models (MLLMs) and promptable segmentation models.

\begin{figure*}[t]
    \centering
    \includegraphics[width=\textwidth]{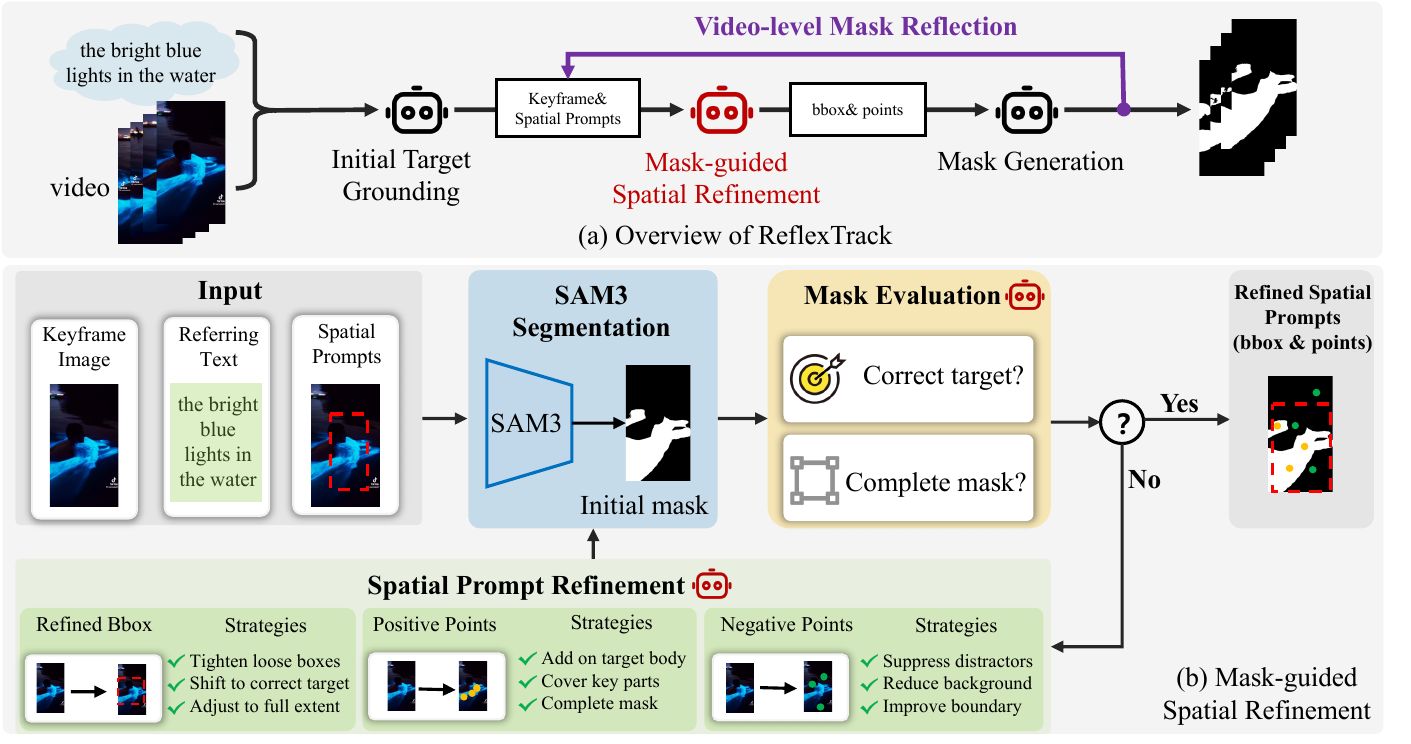}
    \caption{(a) Overview of \method. The primary pipeline performs Initial Target Grounding, Mask-guided Spatial Refinement, and mask generation, while Video-level Mask Reflection is further integrated to identify and repair unreliable predictions. By combining spatial prompt refinement with video-level reflection, \method produces robust RVOS predictions. (b) Illustration of Mask-guided Spatial Refinement. Specifically, SAM~3 first generates an initial mask from the keyframe, referring expression, and spatial prompts. The mask is then evaluated for target correctness and completeness; when an error is detected, the spatial prompts are refined by updating the bounding box and positive and negative points for the next round of segmentation.}
    \label{fig:overview}
\end{figure*}

\begin{figure*}[t]
    \centering
    \includegraphics[width=\textwidth]{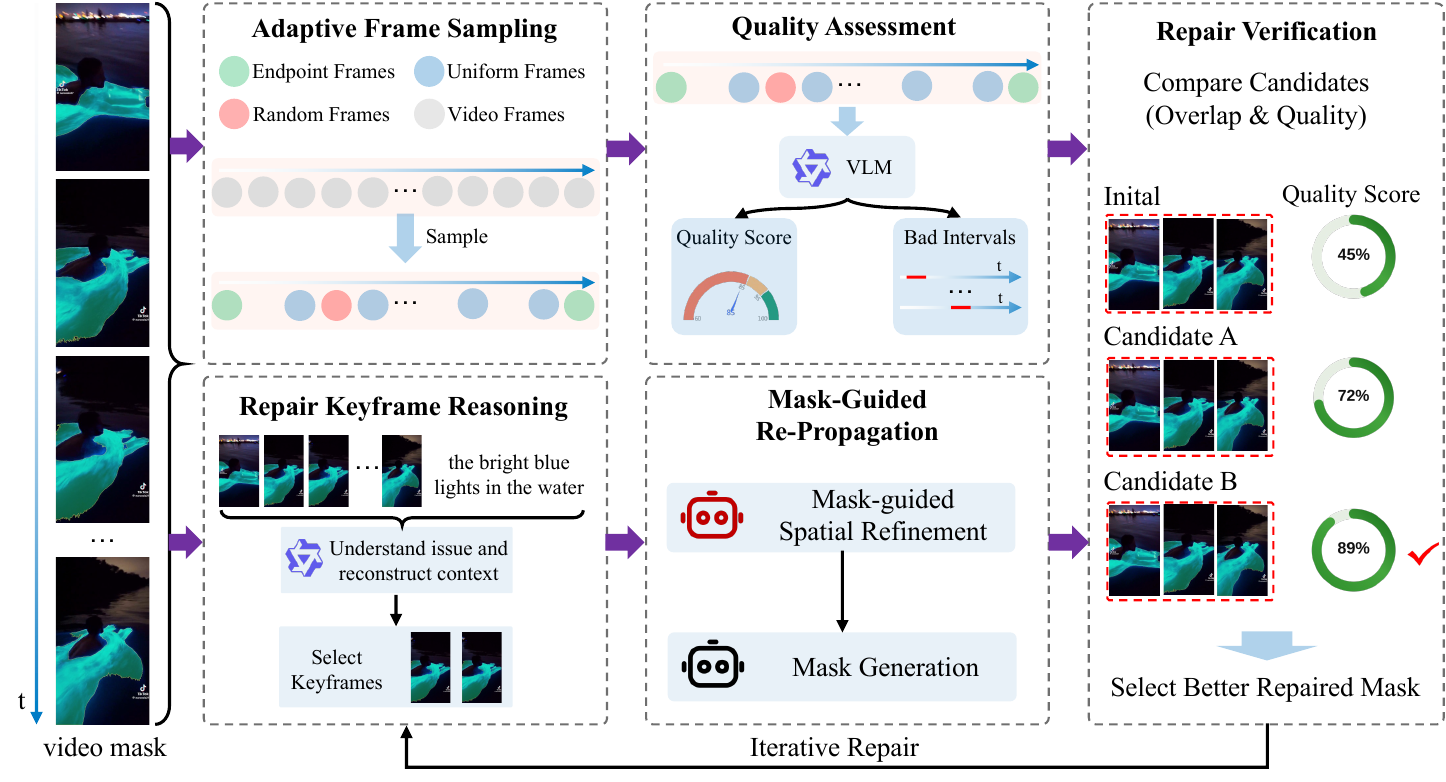}
    \caption{Overview of Video-level Mask Reflection. Adaptive Frame Sampling selects endpoint, uniform, and random frames for efficient sequence-level quality assessment and unreliable-interval localization. Guided by the assessment, the repair agent reconstructs the video context and selects informative keyframes from the full sequence. Mask-guided Spatial Refinement and SAM~3 propagation then generate complementary repair candidates, which are compared with the initial prediction during Repair Verification. The better prediction is retained and can be iteratively refined.}
    \label{fig:reflection}
\end{figure*}

Training-free RVOS provides a flexible alternative by composing pretrained MLLMs, grounding models, and video segmentation foundation models into modular inference pipelines~\cite{huang2024alrefsam2,kao2025cotrvs,jiang2026referagent}. Existing agentic systems can select representative frames, interpret the query, identify the target, and verify semantic consistency without task-specific training. However, their reflection is largely confined to reasoning before dense prediction. The resulting spatial prompts and propagated masks are usually accepted without explicit prediction-level verification.

This open-loop design is fragile. A semantically correct bounding box may still truncate the target, include distractors, or provide weak foreground-background constraints. Likewise, a plausible keyframe mask may drift or disappear during propagation because of occlusion, deformation, fast motion, or temporary target absence. Although SAM~2 and SAM~3 offer strong generic segmentation and propagation capabilities~\cite{ravi2024sam2,carion2025sam3}, their outputs remain sensitive to prompt quality and temporal initialization. Local errors can therefore accumulate into mask loss, target drift, and temporal inconsistency.

We address this limitation with \method, a feedback-driven agent for training-free RVOS. Instead of the one-way process of \emph{reasoning--grounding--segmentation}, \method introduces a closed loop of \emph{reasoning--grounding--segmentation--evaluation--repair}. Feedback operates at two complementary prediction levels.

First, \emph{Mask-guided Spatial Refinement} evaluates the keyframe mask induced by the current spatial prompt. An MLLM checks target correctness and mask completeness, diagnoses spatial errors, and revises the bounding box together with positive and negative point prompts. SAM~3 then regenerates the mask, enabling iterative prompt correction directly guided by downstream segmentation quality.

Second, \emph{Video-level Mask Reflection} inspects the complete mask sequence, localizes low-quality temporal intervals, and reconstructs a more discriminative repair context. It reselects multiple keyframes from the full video, generates complementary candidate sequences through mask-guided re-propagation, and replaces an interval only when repair verification confirms a reliable improvement. This selective strategy preserves high-confidence predictions while focusing expensive MLLM reasoning on problematic regions.

All components remain frozen, requiring no task-specific training, additional pixel-level annotations, or parameter updates. \method achieves an overall $\mathcal{Q}$ score of $69.7$ on Ref-VPS and a $\mathcal{J}\&\mathcal{F}$ score of $67.2$ on ReasonVOS. These results demonstrate that extending reflection from semantic reasoning to spatial and temporal predictions substantially improves the reliability of training-free RVOS.

Overall, our contributions are summarized as follows:
\begin{itemize}[leftmargin=*,topsep=2pt,itemsep=1pt,parsep=0pt]
    \item We propose \method, a training-free RVOS agent that introduces closed-loop feedback over both spatial prompts and video-level mask predictions.
    \item We develop Mask-guided Spatial Refinement, which iteratively revises bounding-box and point prompts using the induced keyframe mask as direct feedback.
    \item We develop Video-level Mask Reflection, which localizes unreliable intervals and selectively repairs them through context reconstruction, keyframe reselection, re-propagation, and verification.
\end{itemize}

\section{Method}
\label{sec:method}

\subsection{Problem Formulation and Overview}
\label{sec:overview}

Given a video $\mathcal{V}=\{I_1,\ldots,I_T\}$ and a natural-language query $q$, RVOS predicts a binary mask sequence $\mathcal{M}=\{M_1,\ldots,M_T\}$ for the referred target, where $M_t\in\{0,1\}^{H\times W}$. \method performs no task-specific training and coordinates frozen MLLMs, a visual grounding module, and SAM~3 for grounding, prompt refinement, mask propagation, and prediction repair.

As shown in Fig.~\ref{fig:overview}, the framework contains four stages. \emph{Initial Target Grounding} selects representative keyframes and produces initial spatial prompts. \emph{Mask-guided Spatial Refinement} evaluates the mask generated from each prompt and iteratively revises its bounding box and positive/negative points. SAM~3 then generates and propagates the target masks to obtain an initial sequence. Finally, \emph{Video-level Mask Reflection} identifies low-quality intervals and selectively repairs them through keyframe reasoning, mask-guided re-propagation, and repair verification. Thus, keyframe masks provide spatial feedback, while sequence-level assessment provides temporal feedback.

\subsection{Mask-guided Spatial Refinement}
\label{sec:spatial_refinement}

A semantically correct grounding result may still be unsuitable for pixel-level segmentation: a loose box introduces background or distractors, whereas a tight box truncates the target. We therefore assess the mask induced by the current prompt and use its errors to refine the prompt itself.

\subsubsection{Mask Generation and Evaluation}
For a selected keyframe $I_k$, let $P_k^{(0)}$ denote the initial prompt, typically a bounding box. At iteration $r$, SAM~3 generates
\begin{equation}
    M_k^{(r)} = \mathcal{S}\!\left(I_k,P_k^{(r)}\right),
    \label{eq:mask_generation}
\end{equation}
where $\mathcal{S}$ is the promptable segmentation model. A mask-evaluation agent jointly observes $I_k$, $q$, $P_k^{(r)}$, and $M_k^{(r)}$ to determine whether the mask selects the correct target and covers it completely. It also diagnoses truncation, background leakage, distractor attachment, and boundary errors, producing feedback $z_k^{(r)}$.

\subsubsection{Prompt Refinement and Propagation}
We represent a prompt as $P_k=\{b_k,P_k^+,P_k^-\}$, containing a bounding box and positive/negative point sets. If the current mask fails evaluation, the refinement agent updates the prompt as
\begin{equation}
    P_k^{(r+1)} = \mathcal{R}_{\mathrm{sp}}\!\left(
    I_k,q,P_k^{(r)},M_k^{(r)},z_k^{(r)}\right),
    \label{eq:prompt_refinement}
\end{equation}
where $\mathcal{R}_{\mathrm{sp}}$ denotes the spatial-refinement agent. The box is tightened, shifted, or expanded according to the diagnosed error; positive points reinforce the target body or missing regions; negative points suppress background and similar distractors. The three prompt types are selected jointly rather than applied in a fixed pattern.

The updated prompt is returned to SAM~3, and the evaluation-refinement loop continues until the mask is judged correct and sufficiently complete or a maximum iteration count is reached. The final prompt $P_k^*$ initializes SAM~3 video propagation, yielding the initial mask sequence $\mathcal{M}^{(0)}$. This procedure corrects spatial initialization errors without model updates.

\subsection{Video-level Mask Reflection}
\label{sec:mask_reflection}

Even with a reliable keyframe initialization, propagation may fail under occlusion, target reappearance, fast motion, or large deformation. Video-level Mask Reflection detects unreliable intervals, generates complementary repair candidates, and retains only verified improvements. As illustrated in Fig.~\ref{fig:reflection}, it comprises Adaptive Frame Sampling, Quality Assessment, Repair Keyframe Reasoning, Mask-Guided Re-Propagation, and Repair Verification.

\subsubsection{Adaptive Frame Sampling and Assessment}
Evaluating every frame with a VLM is expensive and redundant. We construct
\begin{equation}
    \mathcal{S}_{\mathrm{eval}} =
    \mathcal{S}_{\mathrm{end}} \cup
    \mathcal{S}_{\mathrm{uni}} \cup
    \mathcal{S}_{\mathrm{rand}},
    \label{eq:adaptive_sampling}
\end{equation}
where endpoint frames inspect temporal boundaries, uniform frames cover the full sequence with spacing adapted to video length, and random frames reduce the chance of missing short failures. The corresponding image-mask pairs and query are passed to a quality-assessment agent, which evaluates target consistency, mask completeness, false positives during target absence, and temporal coherence. It outputs a sequence-level score and low-quality intervals $\mathcal{B}=\{B_j\}_{j=1}^{J}$.

\subsubsection{Repair Keyframe Reasoning and Re-Propagation}
For each interval, the repair agent analyzes the failure type and reconstructs a more discriminative textual context. It then selects multiple repair keyframes from the complete video, rather than restricting selection to the erroneous interval, so clearer evidence outside the interval can provide stable re-initialization.

For each selected keyframe, the reconstructed context produces an initial spatial prompt, which is further processed by Mask-guided Spatial Refinement. SAM~3 generates the refined keyframe mask and propagates it bidirectionally, producing $C$ complete candidate sequences $\{\mathcal{M}^{(c)}\}_{c=1}^{C}$. Different temporal starting points provide complementary recovery behavior around occlusion, reappearance, and drift.

\subsubsection{Repair Verification and Selective Update}
Repair Verification compares the initial prediction $\mathcal{M}^{(0)}$ with all repaired candidates. For each interval $B_j$, it considers candidate quality---target correctness, mask completeness, and temporal stability---together with overlap consistency among candidates and reliable surrounding frames. The selected candidate is
\begin{equation}
    c_j^* = \argmaxop_{c\in\{0,\ldots,C\}}
    \mathcal{E}\!\left(\mathcal{M}^{(c)}\mid B_j,q\right),
    \label{eq:candidate_selection}
\end{equation}
where $c=0$ denotes the initial prediction. Including the initial sequence prevents forced replacement when no repair is better. The final prediction is updated by
\begin{equation}
\widehat{M}_t =
\begin{cases}
M_t^{(c_j^*)}, & t\in B_j \ \text{and}\
\mathcal{E}(\mathcal{M}^{(c_j^*)})>
\mathcal{E}(\mathcal{M}^{(0)})+\delta,\\
M_t^{(0)}, & \text{otherwise},
\end{cases}
\label{eq:selective_update}
\end{equation}
where $\delta$ is the minimum accepted gain. Only verified intervals are replaced, preserving high-confidence predictions elsewhere. The updated sequence is reassessed, and the loop stops when no reliable improvement remains, the quality requirement is met, or the maximum number of rounds is reached.

\section{Experiments}
\label{sec:experiments}

\subsection{Experimental Setup}
\paragraph{Datasets and metrics.}
We evaluate \method on the new Ref-VPS benchmark and ReasonVOS. For Ref-VPS, we report the overall quality score $\mathcal{Q}$, where a higher value indicates better performance. For ReasonVOS, following Refer-Agent~\cite{jiang2026referagent}, we report the standard $\mathcal{J}\&\mathcal{F}$ score, region similarity $\mathcal{J}$, and contour accuracy $\mathcal{F}$.

\paragraph{Implementation details.}
All experiments are conducted on a single NVIDIA RTX~3090 GPU. The complete pipeline is training-free: Initial Target Grounding, the MLLM-based evaluation and repair agents, and SAM~3 remain frozen throughout inference. For the Ref-VPS evaluation, we report variants using Qwen3.7-Plus and GPT-5.5 as the reflection and repair-verification model. For ReasonVOS, we use Qwen-3.5 9B.

\subsection{Main Results}
\paragraph{Results on Ref-VPS.}
Table~\ref{tab:refvps} compares \method with representative RVS baselines on the new Ref-VPS benchmark. All methods are evaluated using the same overall quality metric $\mathcal{Q}$. The Qwen3.7-Plus variant of ReflexTrack obtains $\mathcal{Q}=64.7$, while replacing the reflection model with GPT-5.5 further improves the score to $69.7$. These results show that the proposed feedback-driven pipeline substantially improves over existing approaches and can directly benefit from stronger MLLM reasoning.

\begin{table}[t]
    \centering
    \captionsetup{font=footnotesize}
    \caption{Comparison on the new Ref-VPS benchmark. All methods are evaluated by the overall quality score $\mathcal{Q}$. The best result is highlighted in bold.}
    \label{tab:refvps}
    \small
    \begin{tabular}{lc}
        \toprule
        Method & $\mathcal{Q}\uparrow$ \\
        \midrule
        MUTR~\cite{yan2024mutr} & 25.4 \\
        UNINEXT~\cite{yan2023uninext} & 28.7 \\
        VD-IT~\cite{zhu2024vdit} & 37.9 \\
        GLUS~\cite{lin2025glus} & 34.6 \\
        REM (MS-1.4B)~\cite{bagchi2025refereverything} & 49.0 \\
        REM (Wan-14B)~\cite{bagchi2025refereverything} & 50.0 \\
        \midrule
        ReflexTrack (Qwen3.7-Plus) & 64.7 \\
        ReflexTrack (GPT-5.5) & \textbf{69.7} \\
        \bottomrule
    \end{tabular}
\end{table}

\paragraph{Results on ReasonVOS.}
Table~\ref{tab:reasonvos} compares ReflexTrack with supervised and zero-shot methods using the results summarized by Refer-Agent~\cite{jiang2026referagent}. ReflexTrack with Qwen-3.5 9B achieves $67.2$ $\mathcal{J}\&\mathcal{F}$, including $70.1$ $\mathcal{J}$ and $64.2$ $\mathcal{F}$. The strong region-similarity score indicates that mask-guided spatial refinement improves target coverage and region accuracy on reasoning-intensive queries.

\begin{table}[t]
    \centering
    \caption{Comparison on ReasonVOS. The best result in each column is highlighted in bold.}
    \label{tab:reasonvos}
    \resizebox{\columnwidth}{!}{%
    \begin{tabular}{lccc}
        \toprule
        Method & $\mathcal{J}\&\mathcal{F}\uparrow$ & $\mathcal{J}\uparrow$ & $\mathcal{F}\uparrow$ \\
        \midrule
        VideoLISA~\cite{bai2024videolisa} & 47.5 & 45.1 & 49.9 \\
        GLUS~\cite{lin2025glus} & 49.9 & 47.5 & 52.4 \\
        RGA3~\cite{wang2025rga3} & 53.6 & 51.3 & 56.0 \\
        CoT-RVS (GPT-4o)~\cite{kao2025cotrvs} & 65.5 & 62.4 & 68.7 \\
        Refer-Agent~\cite{jiang2026referagent} & \textbf{69.8} & 67.0 & \textbf{72.7} \\
        \midrule
        ReflexTrack (Qwen-3.5 9B) & 67.2 & \textbf{70.1} & 64.2 \\
        \bottomrule
    \end{tabular}}
\end{table}

\begin{figure}[t]
    \centering
    \includegraphics[width=\columnwidth]{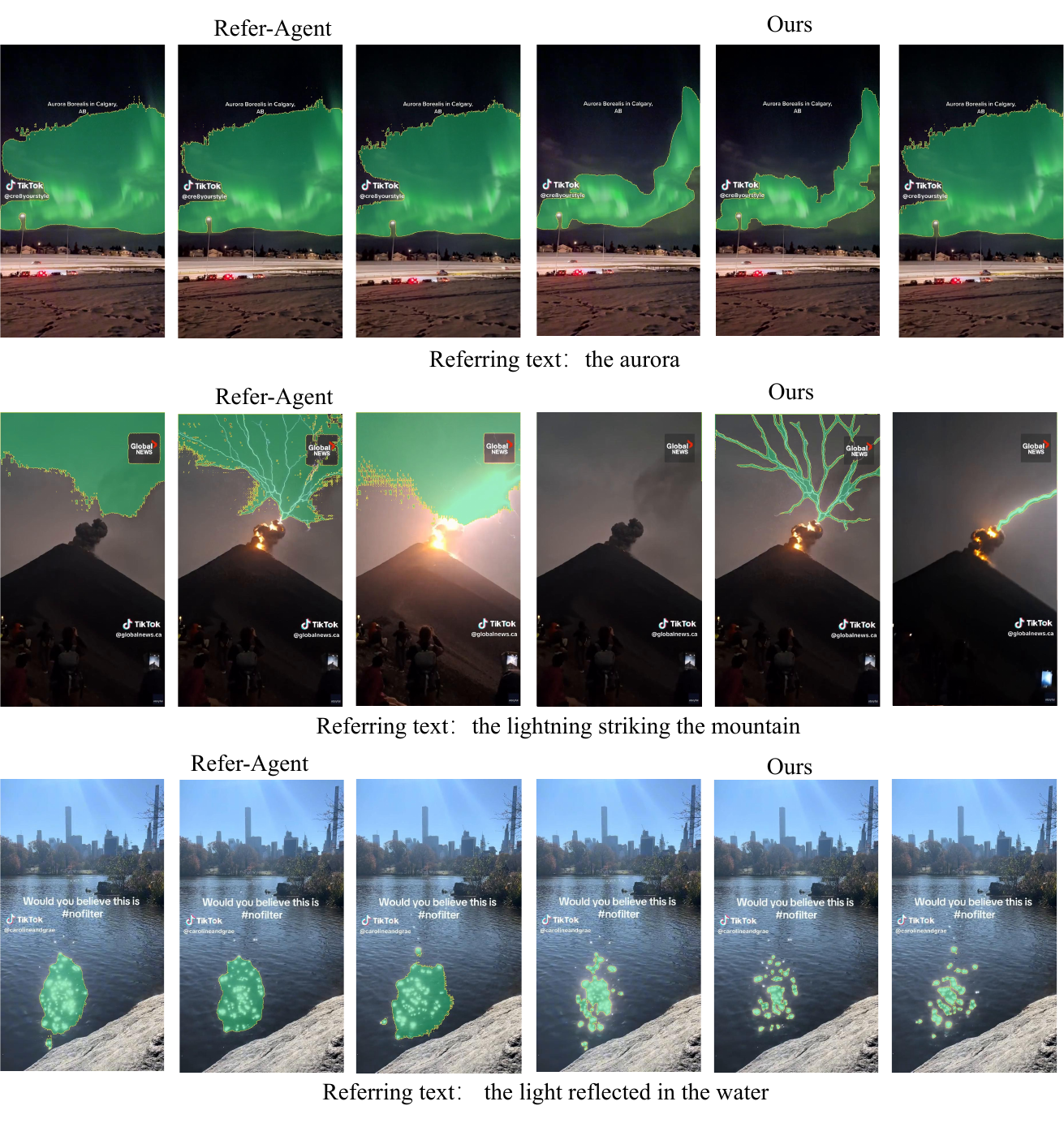}
    \caption{Qualitative comparison with Refer-Agent on Ref-VPS. Each row corresponds to one video, with the first three columns showing Refer-Agent and the last three columns showing ReflexTrack. The examples contain dynamic, ambiguous, and strongly deformable visual concepts.}
    \label{fig:qualitative_comparison}
\end{figure}

\subsection{Ablation Study}
Table~\ref{tab:ablation} isolates the contribution of the two feedback stages and the MLLM used for reflection. The base pipeline obtains $\mathcal{Q}=54.2$. Adding Grounding Refinement increases the score to $60.6$, an absolute gain of $6.4$ points, demonstrating that correcting the initial spatial prompts improves the quality of subsequent mask generation and propagation. Adding Video-level Mask Reflection further raises the score to $64.7$. This gain confirms that spatial refinement and temporal reflection address complementary error sources: the former improves initialization, whereas the latter detects unreliable temporal intervals and selectively replaces them with verified repair candidates.

Replacing Qwen3.7-Plus with GPT-5.5 for reflection and repair verification increases $\mathcal{Q}$ from $64.7$ to $69.7$. Overall, the complete configuration improves over the base pipeline by $15.5$ points, corresponding to a relative gain of approximately $28.6\%$. The result also shows that ReflexTrack can directly exploit stronger MLLM reasoning without task-specific retraining.

\begin{table}[t]
    \centering
    \caption{Ablation study of the feedback components. GR denotes Grounding Refinement.}
    \label{tab:ablation}
    \resizebox{\columnwidth}{!}{%
    \begin{tabular}{lc}
        \toprule
        Method & $\mathcal{Q}\uparrow$ \\
        \midrule
        Base & 54.2 \\
        Base + GR (Qwen3.7-Plus) & 60.6 \\
        Base + GR + Mask Reflection (Qwen3.7-Plus) & 64.7 \\
        Base + GR + Mask Reflection (GPT-5.5) & \textbf{69.7} \\
        \bottomrule
    \end{tabular}}
\end{table}

\subsection{Qualitative Results}
\paragraph{Comparison with Refer-Agent.}
Figure~\ref{fig:qualitative_comparison} compares Refer-Agent with \method on representative Ref-VPS sequences involving aurora, volcanic lightning, and reflected light on water. These examples contain highly deformable and non-object regions whose spatial extent changes substantially over time. Refer-Agent can identify the relevant visual concept, but its masks may become over-expanded, fragmented, or temporally unstable. By refining the initial prompts and repairing unreliable intervals, \method produces masks that more consistently follow the referred dynamic regions across the sequence.

\begin{figure}[t]
    \centering
    \includegraphics[width=\columnwidth]{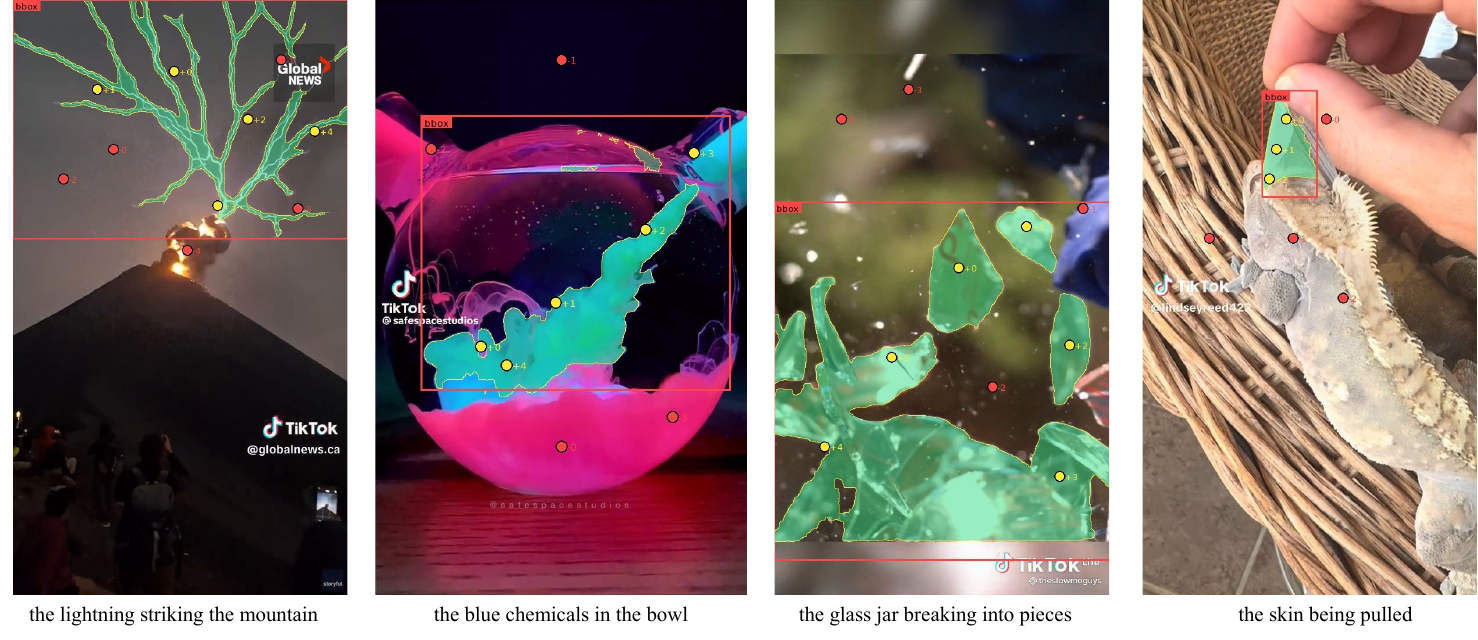}
    \caption{Examples of refined spatial prompts on selected keyframes. Red rectangles indicate refined bounding boxes; yellow and red markers indicate positive and negative point prompts, respectively.}
    \label{fig:spatial_prompts}
\end{figure}

\paragraph{Visualization of refined spatial prompts.}
Figure~\ref{fig:spatial_prompts} visualizes the keyframe prompts produced by Mask-guided Spatial Refinement. Red rectangles denote refined bounding boxes, while yellow and red points represent positive and negative prompts, respectively. Positive points are placed on the target body or missing target regions to strengthen foreground evidence. Negative points are placed on background regions and nearby distractors to suppress false responses. These examples illustrate that the combined box-and-point prompts provide more precise spatial constraints than a single coarse box and therefore offer a more reliable initialization for SAM~3 propagation.

\section{Conclusion}
\label{sec:conclusion}

We introduced ReflexTrack, a training-free, feedback-driven agent for referring video object segmentation. ReflexTrack combines Mask-guided Spatial Refinement, which iteratively improves bounding-box and point prompts using keyframe-mask feedback, with Video-level Mask Reflection, which detects and selectively repairs unreliable temporal intervals. Experiments on Ref-VPS and ReasonVOS demonstrate that these spatial and temporal feedback mechanisms improve segmentation accuracy and robustness without task-specific training or parameter updates.

{\small
\bibliographystyle{ieeenat_fullname}
\bibliography{main}

@inproceedings{gavrilyuk2018actor,
  title     = {Actor and Action Video Segmentation from a Sentence},
  author    = {Gavrilyuk, Kirill and Ghodrati, Amir and Li, Zhenyang and Snoek, Cees G. M.},
  booktitle = {Proceedings of the IEEE/CVF Conference on Computer Vision and Pattern Recognition},
  pages     = {5958--5966},
  year      = {2018}
}

@inproceedings{seo2020urvos,
  title     = {{URVOS}: Unified Referring Video Object Segmentation Network with a Large-Scale Benchmark},
  author    = {Seo, Seonguk and Lee, Joon-Young and Han, Bohyung},
  booktitle = {European Conference on Computer Vision},
  pages     = {208--223},
  year      = {2020}
}

@inproceedings{botach2022mttr,
  title     = {End-to-End Referring Video Object Segmentation with Multimodal Transformers},
  author    = {Botach, Anton and Zheltonozhskii, Evgenii and Baskin, Chaim},
  booktitle = {Proceedings of the IEEE/CVF Conference on Computer Vision and Pattern Recognition},
  pages     = {4985--4995},
  year      = {2022}
}

@inproceedings{wu2022referformer,
  title     = {Language as Queries for Referring Video Object Segmentation},
  author    = {Wu, Jiannan and Jiang, Yi and Sun, Peize and Yuan, Zehuan and Luo, Ping},
  booktitle = {Proceedings of the IEEE/CVF Conference on Computer Vision and Pattern Recognition},
  pages     = {4974--4984},
  year      = {2022}
}

@inproceedings{ding2023mevis,
  title     = {{MeViS}: A Large-Scale Benchmark for Video Segmentation with Motion Expressions},
  author    = {Ding, Henghui and Liu, Chang and He, Shuting and Jiang, Xudong and Loy, Chen Change},
  booktitle = {Proceedings of the IEEE/CVF International Conference on Computer Vision},
  pages     = {2694--2703},
  year      = {2023}
}

@inproceedings{yan2024mutr,
  title     = {Referred by Multi-Modality: A Unified Temporal Transformer for Video Object Segmentation},
  author    = {Yan, Shilin and Zhang, Renrui and Guo, Ziyu and Chen, Wenchao and Zhang, Wei and Li, Hongyang and Qiao, Yu and He, Hao and Gao, Peng},
  booktitle = {Proceedings of the AAAI Conference on Artificial Intelligence},
  volume    = {38},
  number    = {6},
  pages     = {6449--6457},
  year      = {2024}
}

@inproceedings{yan2024visa,
  title     = {{VISA}: Reasoning Video Object Segmentation via Large Language Models},
  author    = {Yan, Cilin and Wang, Haochen and Yan, Shilin and Jiang, Xudong and Li, Yao and Kang, Guoliang and Lu, Weiming and Zhang, Xiangyu and Xie, Jianping},
  booktitle = {European Conference on Computer Vision},
  pages     = {98--115},
  year      = {2024}
}

@inproceedings{lin2025glus,
  title     = {{GLUS}: Global-Local Reasoning Unified into a Single Large Language Model for Video Segmentation},
  author    = {Lin, Lin and Yu, Xiaoyan and Pang, Ziqian and others},
  booktitle = {Proceedings of the IEEE/CVF Conference on Computer Vision and Pattern Recognition},
  year      = {2025}
}

@inproceedings{liang2025referdino,
  title     = {{ReferDINO}: Referring Video Object Segmentation with Visual Grounding Foundations},
  author    = {Liang, Tianming and Lin, Kevin Qinghong and Tan, Chuan and others},
  booktitle = {Proceedings of the IEEE/CVF International Conference on Computer Vision},
  pages     = {20009--20019},
  year      = {2025}
}

@article{huang2024alrefsam2,
  title   = {Unleashing the Temporal-Spatial Reasoning Capacity of {GPT} for Training-Free Audio and Language Referenced Video Object Segmentation},
  author  = {Huang, Shaofei and Ling, Ruyi and Li, Hongyu and others},
  journal = {arXiv preprint arXiv:2408.15876},
  year    = {2024}
}

@article{kao2025cotrvs,
  title   = {{CoT-RVS}: Zero-Shot Chain-of-Thought Reasoning Segmentation for Videos},
  author  = {Kao, Shih-Han and Tai, Yu-Wing and Tang, Chi-Keung},
  journal = {arXiv preprint arXiv:2505.18561},
  year    = {2025}
}

@article{jiang2026referagent,
  title   = {Refer-Agent: A Collaborative Multi-Agent System with Reasoning and Reflection for Referring Video Object Segmentation},
  author  = {Jiang, Haichao and Liang, Tianming and Zheng, Wei-Shi and Hu, Jian-Fang},
  journal = {arXiv preprint arXiv:2602.03595},
  year    = {2026}
}

@inproceedings{ravi2024sam2,
  title     = {{SAM 2}: Segment Anything in Images and Videos},
  author    = {Ravi, Nikhila and Gabeur, Valentin and Hu, Yuan-Ting and others},
  booktitle = {International Conference on Learning Representations},
  year      = {2025}
}

@article{carion2025sam3,
  title   = {{SAM 3}: Segment Anything with Concepts},
  author  = {Carion, Nicolas and Gustafson, Laura and Hu, Yuan-Ting and others},
  journal = {arXiv preprint arXiv:2511.16719},
  year    = {2025}
}

@inproceedings{bagchi2025refereverything,
  title     = {ReferEverything: Towards Segmenting Everything We Can Speak of in Videos},
  author    = {Bagchi, Anurag and Bao, Zhipeng and Wang, Yu-Xiong and Tokmakov, Pavel and Hebert, Martial},
  booktitle = {Proceedings of the IEEE/CVF International Conference on Computer Vision},
  pages     = {23221--23231},
  year      = {2025}
}

@inproceedings{yan2023uninext,
  title     = {Universal Instance Perception as Object Discovery and Retrieval},
  author    = {Yan, Bin and Jiang, Yi and Wu, Jiannan and Wang, Dong and Luo, Ping and Yuan, Zehuan and Lu, Huchuan},
  booktitle = {Proceedings of the IEEE/CVF Conference on Computer Vision and Pattern Recognition},
  year      = {2023}
}

@inproceedings{zhu2024vdit,
  title     = {Exploring Pre-trained Text-to-Video Diffusion Models for Referring Video Object Segmentation},
  author    = {Zhu, Zixin and Feng, Xuelu and Chen, Dongdong and Yuan, Junsong and Qiao, Chunming and Hua, Gang},
  booktitle = {European Conference on Computer Vision},
  year      = {2024}
}

@article{bai2024videolisa,
  title   = {One Token to Seg Them All: Language Instructed Reasoning Segmentation in Videos},
  author  = {Bai, Zechen and He, Tong and Mei, Haiyang and Wang, Pichao and Gao, Ziteng and Chen, Joya and Liu, Lei and Zhang, Zheng and Shou, Mike Zheng},
  journal = {arXiv preprint arXiv:2409.19603},
  year    = {2024}
}

@inproceedings{wang2025rga3,
  title     = {Object-Centric Video Question Answering with Visual Grounding and Referring},
  author    = {Wang, Haochen and Chen, Qirui and Yan, Cilin and Cai, Jiayin and Jiang, Xiaolong and Hu, Yao and Xie, Weidi and Gavves, Efstratios},
  booktitle = {Proceedings of the IEEE/CVF International Conference on Computer Vision},
  pages     = {22274--22284},
  year      = {2025}
}
}

\end{document}